\documentclass[runningheads]{llncs}
\usepackage[T1]{fontenc}
\usepackage{graphicx,verbatim, subfiles}
\usepackage{amsfonts}
\usepackage{algorithm}
\usepackage{algpseudocode}
\usepackage{tikz}
\usepackage{standalone}   
\usepackage{tikz}
\usepackage{amsmath}
\usepackage{graphicx}
\usetikzlibrary{arrows.meta,positioning,calc,fit,backgrounds,shapes.geometric}
\usepackage{fontawesome5}
\usepackage{xcolor}
\usepackage{hyperref}
\usepackage{booktabs}

\hypersetup{
    colorlinks=true,      
    linkcolor=blue,       
    citecolor=green,        
    urlcolor=cyan,        
    pdftitle={My Document},
    pdfauthor={Your Name}
}
\definecolor{vaeBlue}{RGB}{190,220,228}
\definecolor{flowBlue}{RGB}{106,132,150}
\definecolor{clfBlue}{RGB}{20,64,95}
\definecolor{imgGray}{RGB}{210,210,210}

\tikzset{
  arrow/.style={-Latex, line width=1.2pt},
  box/.style={draw, line width=1.2pt, rounded corners=2pt, fill=white, minimum height=9mm, align=center, inner sep=3pt},
  smallbox/.style={box, minimum width=28mm},
  img/.style={draw, line width=1.2pt, fill=imgGray, minimum width=16mm, minimum height=16mm},
}

\begin{document}
\title{Weakly-Supervised Lung Nodule Segmentation via Training-Free Guidance of 3D Rectified Flow}
%

\author{Richard Petersen, Fredrik Kahl and Jennifer Alvén}  
\authorrunning{Anonymized Author et al.}
\institute{Chalmers University of Technology, Gothenburg, Sweden \\
    \email{richard.petersen@chalmers.se}}
  
\maketitle              
\begin{abstract}
Dense annotations, such as segmentation masks, are expensive and time-consuming to obtain, especially for 3D medical images where expert voxel-wise labeling is required. Weakly supervised approaches aim to address this limitation, but often rely on attribution-based methods that struggle to accurately capture small structures such as lung nodules. In this paper, we propose a weakly-supervised segmentation method for lung nodules by combining pretrained state-of-the-art rectified flow and predictor models in a plug-and-play manner. Our approach uses training-free guidance of a 3D rectified flow model, requiring only fine-tuning of the predictor using image-level labels and no retraining of the generative model. The proposed method produces improved-quality segmentations for two separate predictors, consistently detecting lung nodules of varying size and shapes. Experiments on LUNA16 demonstrate improvements over baseline methods, highlighting the potential of generative foundation models as tools for weakly supervised 3D medical image segmentation. 


\end{abstract}

\section{Introduction}
Lung cancer is the deadliest cancer worldwide~\cite{american_cancer}. Early detection of pulmonary nodules through low-dose computed tomography (CT) screening has been shown to reduce mortality by approximately $20\%$ compared to chest radiography~\cite{LUNA16}, yet manual image assessment remains time-consuming and resource-intensive for radiologists~\cite{national2011reduced}. Deep learning-based methods have achieved strong performance in automatic pulmonary nodule detection~\cite{dutande2022deep}, but their training typically relies on large quantities of annotated data. Weakly-supervised segmentation (WSS) offers a potential strategy to address this limitation by learning from weaker forms of supervision, such as image-level labels, points or bounding boxes. However, deriving accurate segmentations from weak supervision alone, particularly for small structures such as lung nodules, remains highly challenging. In this work, we propose a plug-and-play method for weakly-supervised lung nodule segmentations by combining a pretrained 3D rectified flow generative model with a weakly-supervised target predictor through training-free guidance.

\paragraph{Related work.} CAM-based methods~\cite{CAM_paper,selvaraju2017grad,wang2020score} are often used to obtain weakly-supervised segmentations (WSS) by highlighting regions that contribute most to a classification network's prediction, but they tend to emphasize only the most discriminative parts, leading to low-quality segmentation masks~\cite{wang2025weakmedsam}. Reconstruction-based anomaly detection is another common strategy for medical WSS, where variants of autoencoders, generative adversarial networks (GANs)~\cite{schlegl2019f,di2019survey,wolleb2020descargan}, and diffusion models~\cite{mousakhan2024anomaly,sanchez2022healthy,xing2023diff,wyatt2022anoddpm} are trained to reconstruct normal images, with anomalies inferred from reconstruction errors. Guided diffusion has been explored for reconstruction-based anomaly detection in 2D medical imaging~\cite{wolleb2022diffusion}, but prior work requires training both a diffusion model and a noise-dependent classifier from scratch, which limits generalization to new imaging domains without retraining both components. Moreover, achieving good performance requires a large number of sampling steps during guided generation for each 2D slice. 

As an alternative, rectified flow models~\cite{rectified-flow} provide a deterministic formulation that enables substantially faster generation while preserving high-quality results. Latent rectified flow models such as MAISI-v2~\cite{zhao2025maisi} pretrained on CT volumes have demonstrated fast inference and high image quality across diverse anatomies and resolutions compared to diffusion-based counterparts~\cite{guo2025maisi,wang20253d,xu2024medsyn}. MAISI-v2 supports both unconditional and conditional 3D CT image generation by integrating ControlNet~\cite{zhang2023adding} to condition on segmentation masks. However, such conditional generation introduces important limitations: it relies on dense annotations, and extending the model to new conditioning signal requires additional retraining.
The framework of training-free guidance (TFG) enables guiding an off-the-shelf unconditional generative model using a pretrained differentiable target predictor~\cite{bansal2023universal,ye2024tfg,yu2023freedom,patel2025flowchef}. Unlike classifier-guidance~\cite{classifier_guidance}, where the predictor needs to be trained on noisy samples, the target predictor in TFG is trained only on clean samples, thereby avoiding expensive retraining when new target properties are introduced. This opens the possibility of combining pretrained generative models and predictors, which is particularly appealing in medical imaging settings with limited annotations.

\paragraph{Contribution.} We propose a plug-and-play framework for weakly-supervised segmentation in CT volumes by combining a pretrained 3D rectified flow generative model with a weakly-supervised predictor via training-free guidance. Unlike prior counterfactual diffusion and anomaly detection approaches, which require retraining or fine-tuning the generative model or training auxiliary noise-conditioned classifiers, our method operates directly on an off-the-shelf generative model and requires only a differentiable predictor trained with image-level labels. This enables counterfactual generation without modifying the generative model, allowing scalable reuse of large pretrained rectified flow models. Segmentation masks are obtained by comparing the original and guided reconstructions, resulting in improved segmentation agreement compared to attribution-based weakly-supervised methods. The method operates in 3D, preserving volumetric anatomical consistencies and avoiding structural artifacts that commonly arise in slice-wise 2D approaches.

\section{Method}
\begin{figure}[t]
  \centering
  \scalebox{0.9}{\input{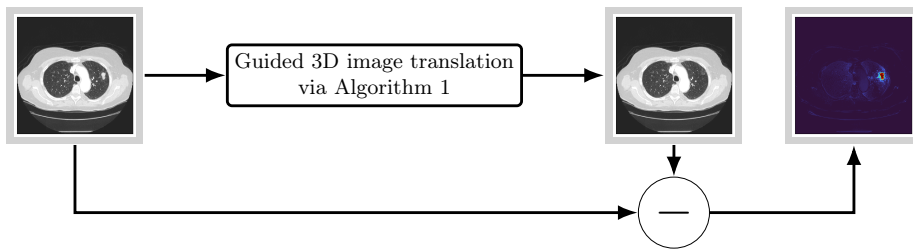}}  
  \caption{
 Overview of the weakly-supervised segmentation (WSS) framework. A predictor-guided rectified flow model generates a counterfactual reconstruction, and the residual image with respect to the input yields the segmentation mask.}
  \label{fig:wss_framework}
\end{figure}

Our method leverages pretrained foundation models in a plug-and-play manner to extract weakly-supervised segmentations of lung nodules. Specifically, we combine MAISI-v2, a state-of-the-art 3D rectified flow model for medical image synthesis, with two alternative predictor models pretrained on large-scale medical imaging data; MedSAM \cite{ma2024segment} and RadImgNet \cite{mei2022radimagenet}. The predictor model is used to guide the generative sampling process towards a counterfactual image corresponding to the absence of lung nodules. Concretely, given a CT volume, we steer the generative trajectory such that the predicted probability of nodule presence is reduced. The weak segmentation mask is then obtained by computing the voxel-wise absolute difference between the original image and the guided counterfactual sample. An overview of the framework is shown in Fig.~\ref{fig:wss_framework}.

\paragraph{Rectified flow.} 
Rectified flow learns a transport map between a source distribution $\pi_0$ and a target distribution $\pi_1$~\cite{rectified-flow}. The model parameterizes a time-dependent vector field $v_\theta(X_t, t)$, represented by a neural network with learnable parameters $\theta$, which transforms samples $X_0 \sim  \pi_0$ into $X_1 \sim \pi_1$ by solving the ordinary differential equation (ODE): 
\begin{equation} \label{ode}
    \mathrm{d}X_t = v_\theta(X_t, t)\, \mathrm{d}t.
\end{equation}

The vector field $v_\theta(X_t, t)$ is learned by minimizing the least squares regression objective:
\begin{equation}\label{fm-loss}
    \mathcal{L \left ( \theta \right )} =  \left\|  (X_1 - X_0) - v_\theta(X_t, t)  \right\|^{2},
\end{equation} 
where $X_t = tX_1 + (1-t)X_0$. This formulation encourages 
linear flows, enabling
high quality results with few sampling steps when solving the ODE in Eq.~\ref{ode}. 

In practice, rectified flow can be performed in a learned latent space using an autoencoder that maps images $X \in \mathbb{R}^{H\times W\times D}$ to latent representations $z \in \mathbb{R}^{h\times w\times d}$. The rectified flow is then applied in the lower-dimensional latent space, resulting in improved scalability for high-dimensional data~\cite{esser2024scaling}.

\begin{figure}[h]
  \centering
  \scalebox{0.65}{\input{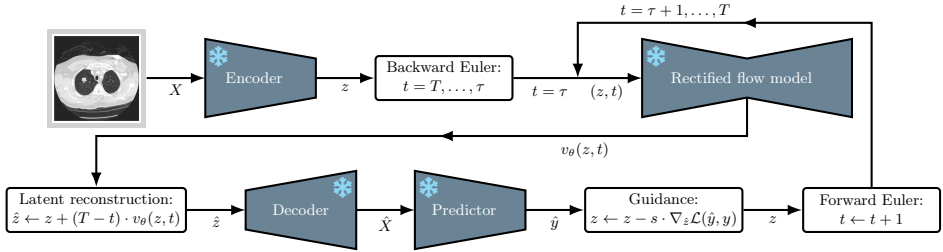}}  
  \caption{Overview of the proposed training-free guidance (TFG) framework for predictor-guided rectified flow in latent space, performed at inference. The symbol \small{\textcolor{cyan!40!white}{\faSnowflake}} indicates that the models are frozen.}
  \label{fig:tfg_framework}
\end{figure}

\paragraph{Training-free guidance.}
In order to avoid costly retraining of the generative model, we leverage the TFG framework \cite{bansal2023universal,ye2024tfg,yu2023freedom}, which enables guiding an arbitrary generative model using a predictor model, rather than training a new conditional model from scratch. Since off-the-shelf medical image classifiers are typically insufficient, we fine-tune a pretrained backbone to serve as a target predictor.  

We guide the unconditional MAISI-v2 rectified flow model using a guidance strategy inspired by FlowChef~\cite{patel2025flowchef}. A brief outline of the method is provided below; see also Algorithm~\ref{alg:tfg} and Fig.~\ref{fig:tfg_framework} for an overview. We omit all time step indices for brevity. First, instead of starting from pure noise, a CT volume $X$ is encoded into a lower dimensional latent representation $z$ using the variational encoder $\mathcal{E}$. The latent representation $z$ is then perturbed using the backward Euler method, Eq. \ref{ode}, to a predetermined intermediate time step $\tau$, in order to preserve anatomical structures during reconstruction. A clean latent estimate is computed as 
\begin{equation}\label{clean_latent_estimation}
     \hat{z} \gets z +  v_\theta(z, t) \cdot (1-t),
\end{equation}
which is then decoded by the variational decoder $\mathcal{D}$ to obtain a reconstruction $\hat{X}$ in image space. This allows us to use the reconstruction $\hat{X}$ as input to the target predictor, yielding predictions $\hat{y}$ used to compute the loss $\mathcal{L}(\hat{y}, y)$, where $y$ denotes the guiding label and $\mathcal{L}$ is the binary-cross entropy loss. The intermediate latent variable is then guided via the gradient update
\begin{equation}\label{guidance}
     z \gets z - s \cdot \nabla_{\hat{z}} \mathcal{L}(\hat{y}, y),
\end{equation}
where $s$ denotes the guidance strength. For in detail explanation of this update, see~\cite{rectified-flow}. The guided latent is subsequently updated according to Eq.~\ref{ode}, and then repeated until the final time step is reached. The weakly-supervised segmentation is finally obtained as the absolute difference between the guided generated image and the original image.  

\begin{algorithm}
\fontsize{8pt}{10pt}\selectfont
\caption{Weak lung nodule segmentation via TFG}\label{alg:tfg}
\begin{algorithmic}
\Statex \textbf{Hyperparameters:} guidance strength $s$, discretization steps $T$, intermediate time step $\tau$, guidance steps $m$
\Statex \textbf{Input:} rectified flow model $v_\theta$, encoder $\mathcal{E}$, decoder $\mathcal{D}$, target predictor $f$, guiding label $y$, loss $\mathcal{L}$, CT volume $X$
\State $z \gets \mathcal{E}(X)$ \Comment{Encode input volume}
\State $dt \gets 1/T$ \Comment{Step size}
\For{$t \in \{T...\tau \}$} \Comment{Noise to intermediate step $\tau$}
\State $z\gets \textsc{BackwardEuler}(v_\theta, z, t, dt)$
\EndFor
\For{$t \in \{\tau...T \}$}
\If{$t < \tau + m$} \Comment{Apply guidance for $m$ steps}
\State $\hat{z} \gets z +  v_\theta(z, t) \cdot (T-t)  \cdot dt$ \Comment{One-step clean latent estimate}
\State $\hat{X} \gets \mathcal{D}(\hat{z})$ \Comment{Decode clean latent estimate}
\State $\hat{y} \gets f(\hat{X})$ \Comment{Compute predictor output}
\State $z \gets z - s \cdot \nabla_{\hat{z}} \mathcal{L}(\hat{y}, y)$ \Comment{Latent guidance update}
\EndIf
\State $z \gets \textsc{ForwardEuler}(v_\theta, z, t, dt)$
\EndFor
\State $X^* \gets \mathcal{D}(z)$ \Comment{Final reconstruction}
\Statex \textbf{Return} $|X^* - X|$ \Comment{Weak segmentation mask} 
\end{algorithmic}
\end{algorithm}


\section{Experiments}
\paragraph{Data and pre-processing.}
We evaluate our method on the LUNA16~\cite{LUNA16} dataset, which consists of 888 thoracic CT scans with annotated lung nodules. LUNA16 is chosen as it provides segmentation masks for quantitative evaluation and since it has not been used for dense segmentation training of MedSAM or RadImgNet. We follow the official 10-fold cross-validation protocol of LUNA16, where in each fold one subset is used for evaluation while the remaining folds are used to fine-tune the predictor models (using image-level labels only). The dense annotations are used solely for evaluation. All CT slices are resized to $256\times 256$, intensities clipped to the Hounsfield Unit range $[-1000,1000]$, and normalized to the $[0,255]$ range to match the expected input format of the pretrained models. 

\subsection{Implementation details} 
\paragraph{Weakly-supervised predictor fine-tuning.} We consider two alternative predictor models with different backbone architectures: the MedSAM TinyViT and RadImgNet ResNet50, due to their large-scale pretraining on medical imaging data. Each backbone is fine-tuned on the training folds using image-level labels only, while the held-out fold is used for evaluation. For MedSAM, the image encoder is adapted for weakly-supervised binary classification by applying adaptive average pooling followed by a $1\times1$ convolutional layer to the last transformer block, following~\cite{wang2025weakmedsam}. For RadImgNet, a linear classification head is added on top of the encoder. We employ a 2.5D strategy by stacking adjacent 2D slices along the channel dimension of the input layer and using the center slice for prediction ~\cite{kumar2024flexible,zhang2019multiple}. In all experiments, 9 adjacent slices are used, based on validation experiments. 
Each slice is assigned class label 1 if the corresponding ground truth segmentation mask contains any lung nodule pixel, and 0 otherwise. To mitigate class imbalance, positive and negative examples are drawn with equal probability during training. The models are fine-tuned for 10,000 iterations using a constant learning rate of $5\times 10^{-4}$. Validation is performed every 100 iterations, and the model weights corresponding to the highest validation F1-score are retained. Standard data augmentations including flipping, rotation, translation, and zooming are applied with probabilities 0.5, 0.5, 1.0 and 0.95, respectively.

\begin{figure}[h]
\label{visual_MEDSAM}
\centering
\includegraphics[width=1.01\textwidth,]{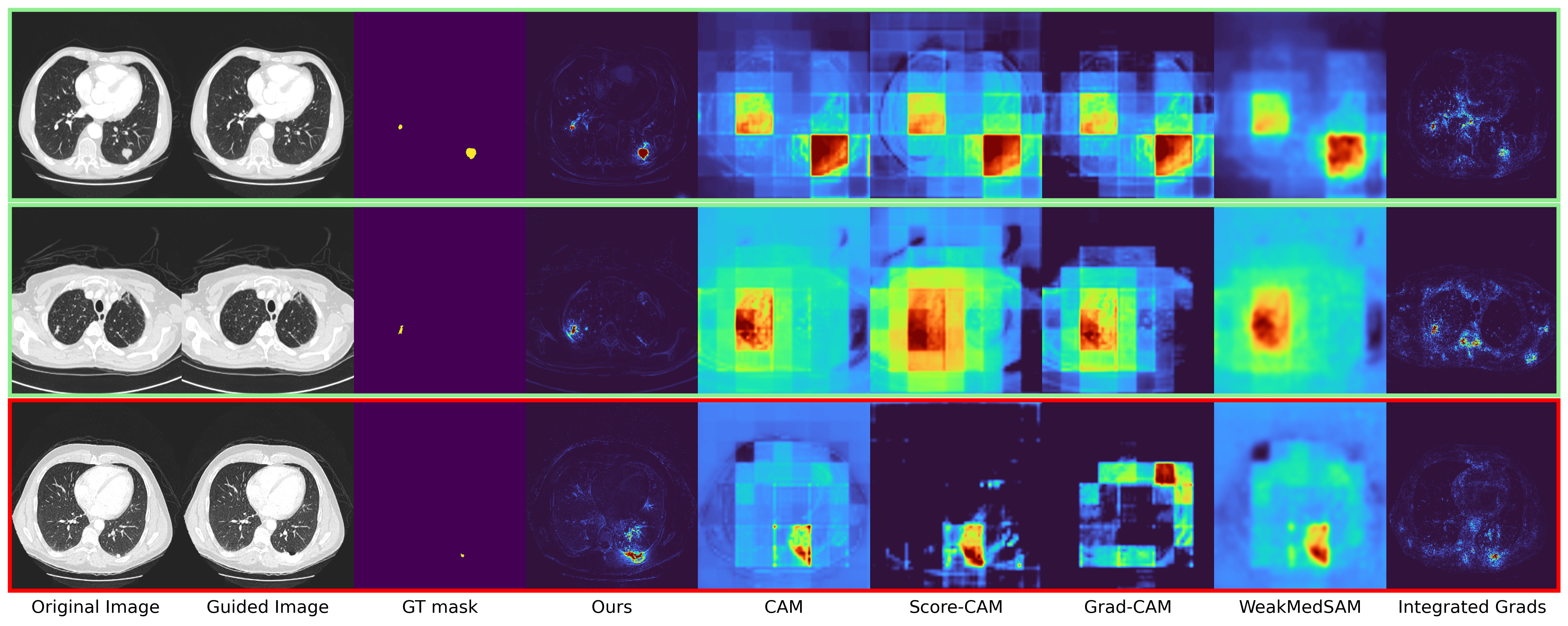}
\caption{Visual comparisons of WSS on LUNA16 for the \underline{MedSAM TinyVit} predictor. Success and failure cases in \textcolor{green!60!white}{green} and \textcolor{red}{red} frame, respectively.  The proposed method suppresses the lung nodules in the guided reconstruction (\textcolor{green!60!white}{Columns 1-2}), resulting in WSS that closely match the shape and size of the ground-truth masks (\textcolor{green!60!white}{Columns 3-4}). The CAM-based methods generally over-segments nodules, producing masks that extend beyond the true lesion boundaries. An example of poor guidance for our method (\textcolor{red}{Columns 2-4}).}
\end{figure}

\paragraph{Training-free guidance segmentation.} During the training-free guidance stage, each CT volume is resized to $256\times 256\times256$ to match the MAISI-v2 encoder. The encoder maps the input image to the latent space, which is subsequently noised to an intermediate time step $\tau =T/2$, where the number of discretization steps $T$ is 30. By Eq.~\ref{clean_latent_estimation} the clean latent is estimated and decoded back to image space. The decoded image is reshaped to match the 2.5D predictor input format described above. Using the decoded image as input to the predictor, the latent representation $z$ is updated according to Eq.~\ref{guidance}, using binary cross-entropy loss and guidance label $y=0$, thereby encouraging suppression of nodule-related features. The guidance loss is computed only for slices predicted to contain nodules, in order to avoid unnecessary computation on slices without nodules. Empirically, the norm of $\nabla_{\hat{z}}$ decreases rapidly during guidance; therefore, guidance is applied only during the first $m=5$ time steps to reduce computational cost. The guidance strength $s$ is set to 1. After the guidance phase, sampling proceeds for the remaining 10 steps using the forward Euler method, resulting in 15 sampling steps in total. The final segmentation mask is obtained by computing the absolute difference between the guided generated image and the original image, followed by thresholding. See Algorithm~\ref{alg:tfg} for a comprehensive overview.

\begin{figure}[h]
\label{visual_RADIMGNET}
\centering
\includegraphics[width=1.01\textwidth,]{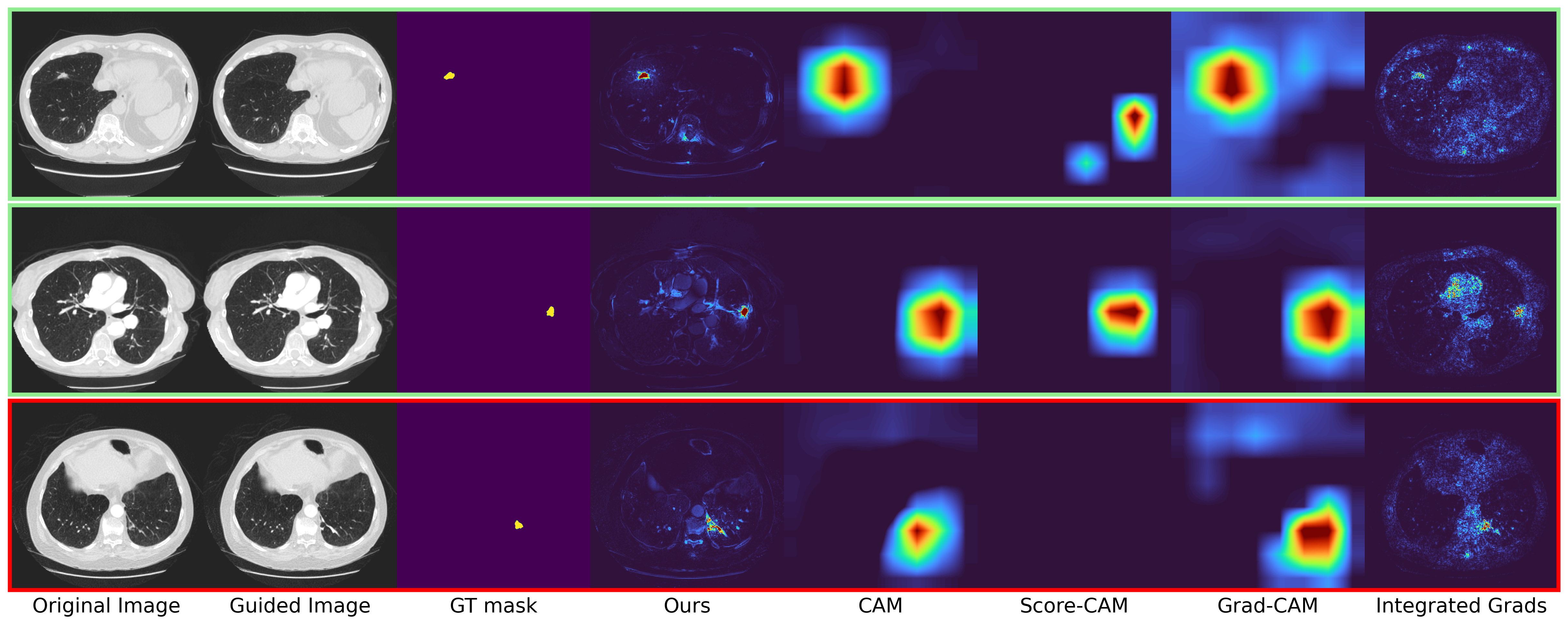}
\caption{Visual comparisons of the WSS on LUNA16 for the \underline{RadImgNet ResNet50} predictor. Similar trends can be observed with a CNN-based predictor, where the proposed method produces masks that more closely follow the ground-truth nodule boundaries compared to the baseline methods.}
\end{figure}

\begin{table}[tb]
\fontsize{10pt}{8pt}\selectfont
\caption{Evaluation on LUNA16 dataset over 10 folds. Metrics with $\downarrow$ indicates lower and $\uparrow$ indicates higher is better, respectively. Best scores are denoted in \textbf{bold} and second best scores are \underline{underlined}. The entries of WeakMedSAM for RadImgNet encoder are empty as its not compatible with the SAM decoder. (Wilcoxon signed-rank test, *$p<0.05$, **$p<0.1$).}\label{tabl_segresults_metrics_LUNA16}
\centering
\begin{tabular}{ l c|c c}
\hline
\multicolumn{1}{l}{Backbone} & \multicolumn{1}{c|}{Method} & \multicolumn{1}{c|}{Mean DSC (\%) $\uparrow$} & \multicolumn{1}{c|}{Median MSD (mm) $\downarrow$} \\
\hline
 & Integrated Grads \cite{sundararajan2017axiomatic} & \underline{36.95}$\pm$\underline{5.05} & 31.72 \\
 & CAM \cite{CAM_paper} & 29.04$\pm$6.77 & 25.84  \\
MedSAM & Grad-CAM \cite{selvaraju2017grad} & 30.88$\pm$7.38 & 28.13 \\
 & Score-CAM \cite{wang2020score} & 30.42$\pm$5.07 & \underline{22.42}  \\
 & WeakMedSAM \cite{wang2025weakmedsam} & 35.07$\pm$4.32  & 73.43  \\
 
 & \textit{Ours} & \textbf{42.05}$\pm$\textbf{4.24*} & \textbf{12.50*}  \\
\midrule
\midrule
 & Integrated Grads \cite{sundararajan2017axiomatic} & \underline{33.89}$\pm$5.20 & 201.87  \\
 & CAM \cite{CAM_paper} & 19.23$\pm$ 5.91 & \underline{44.63}  \\
RadImgNet & Grad-CAM \cite{selvaraju2017grad} & 14.77$\pm$4.21 & 69.41  \\
 & Score-CAM \cite{wang2020score} & 26.19$\pm$3.27 & 83.26  \\
 & WeakMedSAM \cite{wang2025weakmedsam} & - & -  \\
 & \textit{Ours} & \textbf{35.01}$\pm$\textbf{3.63*} & \textbf{44.42**}  \\
\hline
\end{tabular}
\end{table}

\paragraph{Experimental results.}
We evaluate the proposed WSS method in a plug-and-play setting where the predictor is pretrained and kept fixed. For a fair comparison, all methods use the same fine-tuned predictor model. We therefore restrict the comparison to approaches that operate on a trained predictor without requiring comprehensive architectural changes or joint retraining of additional components, such as a generative model. The produced WSS are evaluated using Dice Similarity Coefficient (DSC) and Mean Surface Distance (MSD). 

Table~\ref{tabl_segresults_metrics_LUNA16} summarize the quantitative results on LUNA16 across 10 folds. Using the MedSAM backbone, our method achieves the highest mean DSC ($42.05\%$) and the lowest median MSD (12.50 mm) among all compared methods, indicating better agreement with the size and shape of the nodules. When using the RadImgNet backbone, overall performance decreases across all methods. Nevertheless, our approach achieves the highest DSC ($35.01\%$) and lowest median MSD (44.42 mm). These results indicate that the proposed guidance mechanism generalizes across predictor architectures, although the final segmentation quality remains dependent on the underlying predictor capacity.

Qualitative examples in Fig.~\ref{visual_MEDSAM} support the quantitative findings for the MedSAM predictor. The CAM-based methods are generally capable to localize lung nodules but tend to over-segment, highlighting their limitations in accurately delineating small structures, even when combined with refinement strategies such as WeakMedSAM, which has been reported to achieve state-of-the-art performance among medical WSS methods. Integrated Gradients produces more shape-consistent segmentations, but often under-segments and also introduces false positives. Similar trends can be observed for the RadImgNet predictor in Fig.~\ref{visual_RADIMGNET}. Although overall segmentation quality is lower, our method yields more spatially aligned masks compared to the baselines. Again, this indicates that the proposed framework functions across predictor architecture types, while the final segmentation remains dependent on the underlying predictor capacity.

\section{Conclusion}
In this work, we present a method for extracting pulmonary nodule segmentations from pretrained models in a fully weakly supervised manner, requiring minimal additional effort beyond simple fine-tuning on image-level labels. The proposed method combines off-the-shelf rectified flow and predictor models via the TFG framework, thereby avoiding costly retraining. Our approach demonstrates, both quantitatively and qualitatively, improved segmentation quality with respect to the size and shape compared to commonly used methods for WSS. A key advantage of the proposed framework is the decoupling of generative modeling and downstream task adaptation, which reduces the need for expensive retraining and allows flexible integration with different predictor architectures. By enabling improved lung nodule WSS, the proposed framework offers a practical alternative to manual voxel-wise annotation. 

\subsubsection{Acknowledgment.}
The computations were enabled by resources provided by the National Academic Infrastructure for Supercomputing in Sweden (NAISS), partially funded by the Swedish Research Council through grant 2022-06725. The study was supported by the Sjöberg Foundation grant 2022-489.
%
%
%
%
\bibliographystyle{plain}  
\bibliography{sections/references}
\end{document}